\pdfoutput=1

\documentclass[11pt]{article}

\usepackage{acl}

\usepackage{times}
\usepackage{latexsym}

\usepackage[T1]{fontenc}

\usepackage[utf8]{inputenc}

\usepackage{microtype}

\usepackage{inconsolata}
\usepackage{xcolor}
\usepackage{amsfonts,amsmath,amssymb}
\usepackage{pgfplots}
\pgfplotsset{compat=1.9}
\usepgfplotslibrary{colorbrewer}
\usepackage{tikz}
\usetikzlibrary{positioning}
\usepackage{subcaption}
\usepackage{multirow}

\definecolor{yrbL0}{HTML}{767676}
\definecolor{yrbL1}{HTML}{ffb32c}
\definecolor{yrbL2}{HTML}{f23b35}
\definecolor{yrbL3}{HTML}{297aed}
\definecolor{yrbL4}{HTML}{049e4e}
\definecolor{yrbL5}{rgb}{0.69, 0.61, 0.85}

\newcommand{\squishlist}{
 \begin{list}{$\bullet$}
  { \setlength{\itemsep}{0pt}
     \setlength{\parsep}{3pt}
     \setlength{\topsep}{3pt}
     \setlength{\partopsep}{0pt}
     \setlength{\leftmargin}{1.5em}
     \setlength{\labelwidth}{1em}
     \setlength{\labelsep}{0.5em} } }

\newcommand{\squishlisttwo}{
 \begin{list}{$\bullet$}
  { \setlength{\itemsep}{0pt}
     \setlength{\parsep}{0pt}
    \setlength{\topsep}{0pt}
    \setlength{\partopsep}{0pt}
\setlength{\leftmargin}{2em}
\setlength{\labelwidth}{1.5em}
\setlength{\labelsep}{0.5em} } }

\newcommand{\squishend}{
\end{list}  }

\title{PromptAttack: Probing Dialogue State Trackers with Adversarial Prompts}

\author{Xiangjue Dong$^1$, Yun He$^1$\thanks{\hspace{0.2cm}Equal Contribution}, Ziwei Zhu$^2$\footnotemark[1], James Caverlee$^1$\\
$^1$ Texas A\&M University, $^2$ George Mason University \\ \small\texttt{\{xj.dong, yunhe, caverlee\}@tamu.edu, zzhu20@gmu.edu}}

\begin{document}
\maketitle
\begin{abstract}
A key component of modern conversational systems is the Dialogue State Tracker (or DST), which models a user's goals and needs. Toward building more robust and reliable DSTs, we introduce a prompt-based learning approach to automatically generate effective adversarial examples to probe DST models. Two key characteristics of this approach are: (i) it only needs the output of the DST with no need for model parameters, and (ii) it can learn to generate natural language utterances that can target any DST. Through experiments over state-of-the-art DSTs, the proposed framework leads to the greatest reduction in accuracy and the best attack success rate while maintaining good fluency and a low perturbation ratio. We also show how much the generated adversarial examples can bolster a DST through adversarial training. These results indicate the strength of prompt-based attacks on DSTs and leave open avenues for continued refinement.
\end{abstract}

\section{Introduction}

\begin{figure}[t]
\centering
    \begin{subfigure}{\columnwidth} \centering
   \includegraphics[width=1.03\columnwidth]{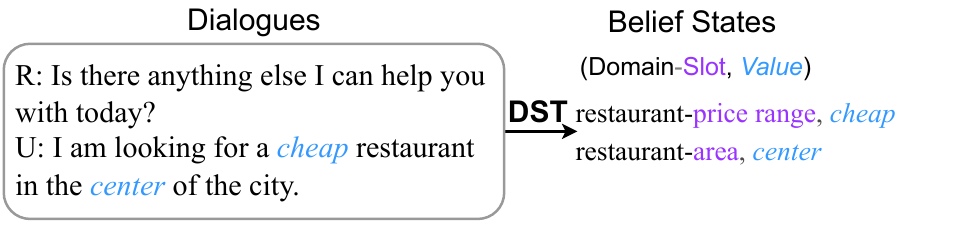}
    \caption{Dialogue state tracking task. R represents system response, and U represents user utterance.}
    \label{fig:non-attack}
    \end{subfigure}
    \hspace{\fill}
    \begin{subfigure}{\columnwidth} \centering
    \includegraphics[width=1.03\columnwidth]{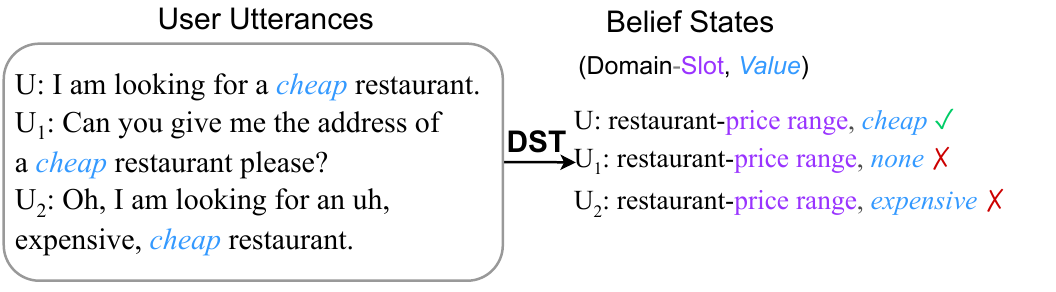}
    \caption{Adversarial examples. U$_{1}$ and U$_{2}$ are adversarial examples based on U which maintain ground-truth values but lead DST models to wrong predictions.}
    \label{fig:attack}
    \end{subfigure}
    \hspace{\fill}
    \vspace{-0.4cm}
 \caption{Dialogue examples and adversarial examples.}    
\label{fig:dialogue}
\vspace{-0.3cm}
\end{figure}

Task-oriented dialogue systems aim to help users with tasks through a natural language conversation. Example tasks include booking a hotel or completing a do-it-yourself project. A key component for enabling a high-quality task-oriented dialogue system is the \textit{Dialogue State Tracker} (or DST) which plays an important role in understanding users' goals and needs~\cite{wu2019transferable,hosseini-etal-2020-simple,li2021coco, dai-etal-2021-preview, feng-etal-2021-sequence,zhao-etal-2021-effective-sequence,balaraman-etal-2021-recent}. For example in Figure~\ref{fig:non-attack}, given the user utterance ``I am looking for a cheap restaurant in the center of the city'', the DST extracts the user's preference for booking a restaurant, which is typically represented as slot-value pairs such as \texttt{(restaurant-price range, cheap)} and \texttt{(restaurant-area, center)}. The current state of the conversation is a primary driver of the subsequent dialogue components (e.g., what is the next action to take? what is the appropriate response to generate?). 

For a conversational system designer, it is critical that a deployed DST be robust and reliable, even in the presence of a wide variety of user utterances. Many of these systems are trained over previous user utterances and so may have only limited coverage of the space of these utterances. Further, beyond these benign users, there is also a long history of spammers, trolls, and malicious users who aim to intentionally undermine deployed systems.

Indeed, recent work has demonstrated that careful construction of adversarial examples can cause failures in the DST~\cite{li2021coco, liu-etal-2021-robustness}, leading to incorrect slot-value pairs and degraded user experience. These approaches, however, are mainly hand-crafted or based on heuristics. As a result, there is a research gap in learning-based methods for probing DSTs centered around three key questions: (i) How can we systematically learn effective adversarial examples? (ii) What impact do such discovered examples have on the quality of state-of-the-art DSTs? and (iii) Can we build more robust DSTs even in the presence of such adversarial examples? Further compounding these questions are the inherent challenges of adversarial examples in the context of a DST: that is, the examples should preserve the semantics of a non-adversarial input while leading to an incorrect prediction \textit{even in the presence of the correct slot-value in the adversarial input} as illustrated in Figure~\ref{fig:attack}. For example, an adversarial example based on the user utterance ``I am looking for a cheap restaurant'' that maps to the slot-value pair \texttt{(restaurant-price range, cheap)} should preserve the user intent for ``cheap'' while leading to the incorrect prediction  \texttt{(restaurant-price range, expensive)}.

Hence, in this paper, we propose a novel prompt-based learning approach called \textit{PromptAttack} to automatically generate effective adversarial examples to probe DST models. Our approach builds on recent advances in prompt learning, which has demonstrated a strong ability in probing knowledge in pre-trained language models for many NLP tasks~\cite{gao-etal-2021-making,li-liang-2021-prefix,liu2021pre,zhu-etal-2022-continual}. Concretely, we first show how to find effective adversarial prompts in both a discrete and a continuous setting. In both cases, our approach needs only the output of the DST (e.g., \texttt{(restaurant-price range, cheap)}) with no need for model parameters or other model details. Second, we use the adversarial prompts to generate adversarial examples via a mask-and-filling protocol, resulting in natural language utterances that can be targeted at any DST. As a result, such a prompt-based attack can be widely applied. 

Through experiments over four state-of-the-art DSTs and versus competitive baselines, we find that the prompt-based framework leads to the greatest reduction in accuracy for all DSTs, ranging from a 9.3 to 31.0 loss of accuracy of the DST making a correct slot-value prediction. Further, we observe that PromptAttack results in the best attack success rate (that is, how many of the adversarial examples lead to incorrect predictions). Moreover, the generated adversarial examples maintain good fluency and low perturbation ratio, evidence that they are close to legitimate non-adversarial user inputs. We also show how such a prompt-based attack can be used to bolster a DST by augmenting the original training data with adversarial examples, leading to a significant increase in accuracy (from 61.3 to 67.3). These and other results indicate the strength of prompt-based attacks on DSTs and leave open avenues for continued refinement.\footnote{Our code is publicly available at \url{https://github.com/dongxiangjue/PromptAttack}.}

\section{Related Work}

Adversarial examples have been widely explored to investigate the robustness of models~\cite{goodfellow2014explaining}. Recent work in the NLP domain has targeted tasks like text classification and inference~\cite{pruthi-etal-2019-combating,ren-etal-2019-generating,morris-etal-2020-textattack,jin-etal-2020-robust,li-etal-2020-bert-attack,yang2022prompting,lei2022phrase}, reading comprehension~\cite{jia2017adversarial,bartolo-etal-2021-improving}, named entity recognition~\cite{simoncini-spanakis-2021-seqattack}, and machine translation~\cite{belinkov2018synthetic}. These works typically aim to construct examples that are imperceptible to human judges while misleading the underlying model to make an incorrect prediction, while also maintaining good fluency and semantic consistency with original inputs~\cite{li-etal-2020-bert-attack}. Only a few works have begun to explore adversarial examples in DSTs like CoCo~\cite{li2021coco}, which aims to test the robustness of models by creating novel and realistic conversation scenarios. They show that DST models are susceptible to both unseen slot values generated from in and out of the slot domain. \citet{liu-etal-2021-robustness} propose a model-agnostic toolkit to test the robustness of task-oriented dialogue systems in terms of three aspects: speech characteristics, language variety, and noise perturbation. The adversarial examples are based on heuristics and it is unclear how to adapt such an approach to new victim models effectively without more hand-crafted templates. In contrast, we explore in this paper the potential of a learning-based approach to generate effective adversarial examples.

\begin{figure*}[htbp!]
\centering
    \begin{subfigure}[t]{0.6\textwidth} \centering
   \includegraphics[width=\linewidth]{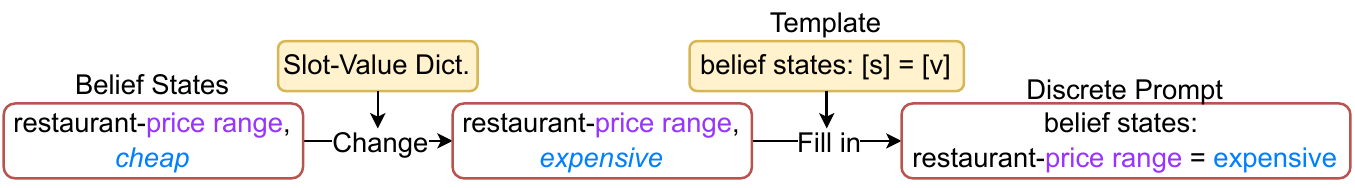}
    \caption{Discrete prompt construction. The discrete prompt is constructed by filling pre-designed templates with slots extracted from the DST model and corresponding random values.}
    \label{fig:discrete}
    \end{subfigure}
    \hspace{\fill}
    \begin{subfigure}[t]{0.38\textwidth} \centering
    \includegraphics[width=0.75\linewidth]{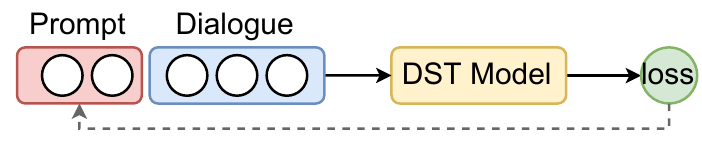}
    \caption{Continuous prompt tuning. The continuous prompt is prepended before the dialogue context embeddings and tuned by optimizing the loss while keeping DST model parameters fixed.}
    \label{fig:continuous}
    \end{subfigure}
    \begin{subfigure}[t]{\textwidth} \centering
    \includegraphics[width=0.5\textwidth]{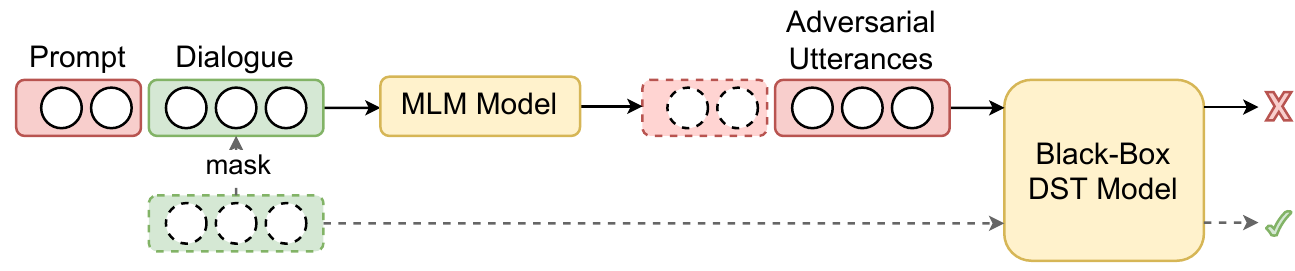}
    \caption{Adversarial example generation. The adversarial prompt (discrete or continuous) is prepended before the masked dialogue context (or embeddings) to generate perturbations via mask-and-filling. After removing the adversarial prompt, the generated adversarial example is used to attack victim models.}
    \label{fig:adversarial}
    \end{subfigure}

 \caption{Overview of PromptAttack.}    
\label{fig:pipeline}
\end{figure*}

Prompt learning is a recently proposed paradigm for using prompts to better probe and adapt large pre-trained language models (PLMs) to a variety of NLP tasks, e.g., text classification and inference~\cite{gao-etal-2021-making,yang2022prompting,wang2022continuous}, factual probing~\cite{zhong-etal-2021-factual}, summarization~\cite{li-liang-2021-prefix}, and dialogue systems~\cite{madotto2021few,lee-etal-2021-dialogue,zhu-etal-2022-continual,yang2022prompt}. With the increase in the size of PLMs, prompt learning has been shown to be parameter-efficient~\cite{liu2021pre,he2022hyperprompt,lu2023hiprompt}. There are two types of prompts: discrete (or hard) prompts and continuous (or soft) prompts. Discrete prompts are human-designed text strings~\cite{brown-etal-2020-language} while continuous prompts are continuous embeddings. Soft prompts proposed by \citet{lester-etal-2021-power} prepend a sequence of continuous vectors to the input, freeze the language model parameters, and then back-propagate the error during tuning. In this paper, we explore both approaches in the design of our prompt-based attack framework. 

Recent works have begun to explore how prompts can be helpful in exposing fundamental flaws in large language models. \citet{yang2022prompting} shows how to manually design prompts to flip the output of a model for classification tasks. However, it is time-consuming to design and find prompts that are most effective to generate adversarial examples capable of successfully attacking victim models. It is an open question how to leverage prompts for uncovering effective adversarial prompts.

\section{PromptAttack}
Our prompt-based learning approach proceeds in two stages. First, our goal is to identify adversarial prompts that can effectively probe a DST to reveal gaps in its robustness. In the second, we use these prompts to create adversarial examples that can attack DSTs successfully while maintaining good fluency. Figure~\ref{fig:pipeline} shows an overview of the proposed approach. In the following, we first formalize DSTs and the problem of probing a DST. Then, we introduce the details of PromptAttack.

\subsection{Task Formulation}
\paragraph{DST Task.} Let $C_{T}=\{(r_{1},u_{1}), \dots, (r_{T}, u_{T})\}$ represent a $T$-turn dialogue, where $r_{i}$ and $u_{i} (1 \leq i \leq T)$ are the system response and user utterance at the $i$-th turn, respectively. Each turn $(r_{i},u_{i})$ contains several slots (e.g., arrive by, leave at) in a specific domain (e.g., taxi), where we denote the $N$ domain-slot pairs as $S=\{s_{1}, \dots, s_{N}\}$. At turn t, we denote current user utterance $u_{t}$ and previous dialogue context $C_{t}=\{(r_{1}, u_{1}), \dots, (r_{t-1}, u_{t-1}), r_{t}\}$. A DST model aims to extract the dialogue belief state $B_{t}=\{(s_{1}, v_{1}), \dots, (s_{N}, v_{N})\}$ for $u_{t}$, where $v_{j}$ is the associated value for each slot $s_{j} (1 \leq j \leq N)$. For example, given a dialogue (\textit{``\dots. I am looking for expensive Mediterranean food.''}), the DST model aims to extract \texttt{expensive} for the slot \texttt{restaurant-price range} and \texttt{Mediterranean} for the slot \texttt{restaurant-food}. 

\paragraph{Attacking a DST.} Given dialogue history $C_{t}$, current user utterance $u_{t}$, and dialogue belief states $B_{t}$, the purpose of an adversarial attack on a DST  is to intentionally perturb the original user utterance $u_{t}$ to get an adversarial example $u_{t}^{'}$ with the two following characteristics: (i) it should mislead the DST model $f$ to incorrectly predict $B_{t}^{'}$, and (ii) it should be fluent in grammar and consistent with the semantics of the original utterance $u_{t}$ by keeping the slot-value-related information in $u_{t}$ unchanged.
If the adversary can achieve $f(u_{t}^{'}) = B_{t}^{'}$, we say the adversarial example $u_{t}^{'}$ attacks $f$ successfully.

\subsection{Finding Adversarial Prompts}
We begin by focusing on the first stage of PromptAttack: how to find the most effective adversarial prompts. We explore both discrete prompts (as illustrated in Figure~\ref{fig:discrete}) and continuous prompts (as illustrated in Figure~\ref{fig:continuous}). A discrete prompt approach is a human-designed natural language prompt that is easy to interpret. We pair this with a treatment of continuous prompts that have more representation capacity.

\paragraph{Discrete Prompt Construction.} To begin with, how can we design discrete prompts? For the DST task, it is time-consuming to manually design sentences containing values that are opposite to the ground truth values for each slot as adversarial prompts. Thus, we apply an intuitive template derived from belief states as an adversarial prompt template: ``belief states: \texttt{[s]} = \texttt{[v]};''. First, we use the DST model to extract value $v_{i}$ for each slot $s_{i}$ in $u_{t}$. If $v_{i}$ is not empty, the corresponding slot name $s_{i}$ is filled in \texttt{[s]}. Then we pick a random value $v_{i}^{'}$ from a predefined in-domain Slot-Value Dictionary~\cite{li2021coco} where $v_{i}^{'}$ and $v_{i}$ are under the same slot $s_{i}$. The new random value $v_{i}^{'}$ is used to fill the \texttt{[v]} in the template. Thus, the adversarial prompt becomes ``belief states: $s_{i}$ = $v_{i}^{'}$;''. As in Figure~\ref{fig:discrete}, given $u_{t}$ (``I am looking for cheap food.''), the predicted $B_{t}$ is \{(\textit{restaurant-price range, cheap})\}, then the adversarial prompt is \textit{``belief states: restaurant-price range = expensive''}, where ``expensive'' is a random value that is different from the predicted value ``cheap''. 

Such a template does not have access to true slot-value pairs of the test set and only utilizes the predictions from the victim models. Since the discrete prompts are human-designed, they are more human-readable and easier to interpret. However, to obtain a prompt for each input, victim models must be queried multiple times, which may be unrealistic in some scenarios.  Hence, we take the next step to search for better prompts in the embedding space of the model. Specifically, we directly optimize the continuous input embedding space through continuous prompt tuning to find the adversarial prompt vectors that are most effective.

\smallskip
\noindent \textbf{Continuous Prompt Tuning.} Continuous prompts are input-agnostic sequences of embeddings with tunable parameters that are optimized directly in the continuous embedding space of the model, as shown in Figure~\ref{fig:continuous}. 
In our task, the length of continuous prompt $\textbf{p}_{att}$ is $m$, denoted as $\textbf{p}_{att}=\textbf{p}_{1}\dots\textbf{p}_{m}$ where each $\textbf{p}_{i}\in \mathbb{R}^{d}(1 \leq i \leq m)$ is a dense vector with the same dimension $d$ as the DST's input embedding (e.g., 768 for TripPy). Given the initialization of $\textbf{p}_{att}$, we concatenate it with the representation of user utterance $\textbf{e}_{u}$ and update it by keeping all other model parameters fixed and optimize the loss of the training set. To find the adversarial prompts $\textbf{p}_{att}$ that could lead DST models $f$ to wrong predictions $B_{t}^{'}$ effectively, we maximize the loss for the ground truth belief states $B_{t}$ for all user utterance in the training set with the following objective:
\[
    \underset{\textbf{p}_{att}}{\arg\max}\,\,\, \mathbb{E}_{\textbf{u}\sim \mathcal{U}}\left[ \mathcal{L}\left( B_{t},f\left( \textbf{p}_{att};\textbf{e}_{u} \right) \right) \right],
    \label{eqa:1}
\]
where $\mathcal{U}$ are user utterances and $\mathcal{L}$ is the loss function of the DST task.
By maximizing the loss for the ground truth belief states we aim to find prompts that force the model to make the most wrong predictions by pushing far apart from the ground truth, like guessing ``expensive'' instead of ``cheap'' for $u_{t}$ (``I am looking for cheap food.'').

In addition, we explore an alternative tuning objective -- minimizing the loss. We replace all the non-empty values in $B_{t}$ to empty (e.g., \texttt{(restaurant-price range, expensive)} changes to \texttt{(restaurant-price range, none)}) and then minimize the loss: 
\[
    \underset{\textbf{p}_{att}}{\arg\min}\,\,\, \mathbb{E}_{\textbf{u}\sim \mathcal{U}}\left[ \mathcal{L}\left( B_{t}^{'},f\left( \textbf{p}_{att};\textbf{e}_{u} \right) \right) \right],
    \label{eqa:2}
\]
where $B_{t}^{'}$ is the set of target belief states. Different from our previous tuning objective, here we aim to find prompts that force the model to fail to extract the correct value for the slot from user utterances. For example, the DST will fail to extract ``cheap'' for slot \textit{price range} in $u_{t}$ (``I am looking for cheap food.'') and thus the predicted belief states will become \texttt{(restaurant-price range, none)}. 

\subsection{Adversarial Example Construction}

Next, we focus on the second stage of PromptAttack: how can we use these prompts to create adversarial examples that can attack DSTs successfully while maintaining good fluency? After obtaining the adversarial prompts, we use them to generate adversarial examples via mask-and-filling~\cite{li-etal-2021-contextualized,yang2022prompting,lei2022phrase} by pre-trained masked language models.  Specifically, we tokenize user utterance $u_{t}$ to a list of tokens, $u_{t}=[w_{u}^{1},w_{u}^{2},\dots,w_{u}^{n}]$. Then we randomly mask tokens that are not values in $B_{t}$, slot-related words, or stopwords with a special token \texttt{[MASK]} and denote the masked $u_{t}$ as $u_{t}^{m}=[w_{u}^{1},$\texttt{[MASK]}$,\dots,w_{u}^{n}]$. 
Shown in Figure~\ref{fig:adversarial}, we concatenate the adversarial prompts and the masked utterance $\textbf{u}_{t}^{m}$ and use a masked language model $\mathcal{M}$ to predict masked text pieces and generate the perturbations based on surrounded context. As shown in Table~\ref{tab:prompt}, for discrete prompt $\textbf{p}_{att}^{d}$, the input for $\mathcal{M}$ would be the concatenation of $\textbf{p}_{att}^{d}$ and $\textbf{u}_{t}^{m}$ while for continuous prompt $\textbf{p}_{att}^{c}$, the input would be the concatenation of $\textbf{p}_{att}^{c}$ and embedding of masked user utterance \textbf{e}$^{1}_{u}$ \texttt{[MASK]} \textbf{e}$^{n}_{u}$. Hence, with  $\textbf{p}_{att}$ and the capability of MLM, the model $\mathcal{M}$ will fill in the blanks with context-consistent tokens which can keep the sentence fluency while maximizing the risk of the DST making wrong predictions, denoted as $P(\texttt{[MASK]}=w | \textbf{p}_{att};\textbf{u}_{t}^{m})$, where $w$ is the generated perturbation. After filling \texttt{[MASK]} with $w$ and removing $\textbf{p}_{att}$, the filled user utterances are used as adversarial examples to attack victim models.
\begin{table}[!tp]
\centering\small
\begin{tabular}{cc}
\hline
\textbf{Method} & \textbf{p}$_{att}$ + u$_{t}^{m}$ (or \textbf{e}$_{u}^{m}$) \\ \hline
PromptAttack$_{d}$ & belief states: \texttt{[s]} = \texttt{[v]}; t$^{1}_{u}$ \texttt{[MASK]} t$^{n}_{u}$ \\
PromptAttack$_{c}$ & \textbf{p}$_{1}$ \textbf{p}$_{2}$ \dots \textbf{p}$_{m}$ $\bigoplus_{}^{}$ \textbf{e}$^{1}_{u}$ \texttt{[MASK]} \textbf{e}$^{n}_{u}$ \\
\hline
\end{tabular}
\caption{Adversarial example generation for discrete prompts and continuous prompts.}
\label{tab:prompt}
\end{table}

\section{Experimental Setup}
Our experiments are designed to test the effectiveness of the proposed prompt-based approach to attack DST models. We structure the experiments around four research questions: \textbf{RQ1}: Are adversarial examples learned by PromptAttack effective and transferable? And how do these examples compare against baseline (non-prompt) approaches? \textbf{RQ2}: Are the generated adversarial examples of good quality? That is, are they fluent with a low perturbation ratio? \textbf{RQ3}: What impact do the design choices of PromptAttack have, i.e., the ratio of perturbed tokens and prompt length? \textbf{RQ4}: And finally, can the generated adversarial examples be used to improve the performance of current DST models to improve their robustness?

\subsection{Dataset}
We evaluate our methods on the widely used and challenging multi-domain dialogue dataset, MultiWOZ 2.1~\cite{eric-etal-2020-multiwoz},\footnote{\url{github.com/budzianowski/multiwoz}, MIT License.} which contains over 10,000 dialogues spanning seven domains. Following existing work~\cite{li2021coco, lee-etal-2021-dialogue, yang2022prompting}, we keep five domains (train, taxi, restaurant, hotel, and attraction) with 30 domain-slot pairs and follow the standard train/validation/test split.

\subsection{Evaluation Metrics}
\label{metrics}
We evaluate the proposed methods with a standard set of metrics~\cite{jin-etal-2020-robust,li-etal-2020-bert-attack,li-etal-2021-contextualized,simoncini-spanakis-2021-seqattack}:
\textbf{Joint goal accuracy (JGA):} the average accuracy of predicting all (domain-slot, value) pairs in a turn correctly. 
\textbf{Attack success rate (ASR):} the proportion of generated adversarial examples that successfully mislead model predictions.
\textbf{Perturbation ratio (PER):} the percentage of perturbed tokens in the sentence. Each replace action accounts for one token perturbed. A lower perturbation ratio indicates more semantic consistency~\cite{li-etal-2020-bert-attack}. 
\textbf{Perplexity (PPL):} a metric to evaluate the fluency of sentences. We calculate the perplexity of adversarial examples through GPT-2~\cite{radford2019language}. PPL is calculated across all the adversarial examples. A lower PPL score indicates higher fluency and naturalness of the adversarial examples.

\subsection{Baseline Methods}
We compare our methods with strong baselines capable of attacking a DST. 
\noindent \textbf{TP} and \textbf{SD} are two methods maintaining the dialogue act labels unchanged and implemented by the LAUG toolkit~\cite{liu-etal-2021-robustness}. For a fair comparison, we do not apply slot value replacement which would modify the slot values in the original utterances.
\noindent \textbf{TP} (Text Paraphrasing) uses SC-GPT~\cite{peng-etal-2020-shot} to generate a new utterance conditioned on the original dialogue acts as data augmentation. 
\noindent \textbf{SD} (Speech Disfluency) mimics the disfluency in spoken language by filling pauses (``um''), repeating the previous word, restarting by prepending a prefix ``I just'' before the original user utterance, and repairing by inserting ``sorry, I mean'' between a random slot value and the original slot value~\cite{liu-etal-2021-robustness}. 

\noindent \textbf{SC-EDA}~\cite{liu-etal-2021-robustness} injects word-level perturbations by synonym replacement, random insertions, swaps, and deletions without changing the true belief states. 
\noindent \textbf{BERT-M} is introduced in this paper as another baseline method. First, we randomly mask tokens that are not slot-value related and not stopwords. Then, we use BERT~\cite{devlin2019bert} to generate perturbations based on the top-$K$ predictions via mask-and-filling,  
where in our experiments $K=20$. We sorted the top 20 tokens based on the possibility scores and pick the one with the lowest possibility to fill the masked position. The filled user utterance is regarded as an adversarial example.

\begin{table*}[!tp]
\centering \small
\begin{tabular}{ccccccc}
\hline
\multirow{2}{*}{\textbf{Method}} & \textbf{TripPy} & \textbf{CoCo} & \textbf{SimpleTOD} & \textbf{TRADE} \\
& \textbf{JGA$\downarrow$ / $\Delta$ / ASR$\uparrow$} & \textbf{JGA$\downarrow$ / $\Delta$ / ASR$\uparrow$} & \textbf{JGA$\downarrow$ / $\Delta$ / ASR$\uparrow$} & \textbf{JGA$\downarrow$ / $\Delta$ / ASR$\uparrow$} \\
\hline
Original   & 61.3 / - / - & 62.6 / - / -& 56.0 / - / - & 49.4 / - / - \\ \hline
SC-EDA  & 60.5 / -0.8 / 1.9 & 61.9 / -0.7 / 1.6 & 53.6 / -2.4 / 9.5 & 48.8 / -0.6 / 4.9\\
TP & 60.3 / -1.0 / 5.6 & 61.5 / -1.1 / 4.7 & 52.6 / -3.4 / 19.3 & 48.8 / -0.6 / 14.1 \\
SD$^{*}$  & 56.5 / -4.8 / 9.3 & 56.1 / -6.5 / 11.4 & 38.8 / -17.2 / 36.6 & \textbf{31.7 / -17.7 / 39.9} \\
BERT-M & 58.9 / -2.4 / 5.0 & 60.1 / -2.5 / 4.8 & 49.6 / -6.4 / 16.4 & 45.9 / -3.5 / 11.5 \\ \hline
PromptAttack$_{d}$ & 53.6 / -7.7 / 16.0 & \underline{53.7} / \underline{-8.9} / \underline{16.9} & 38.9 / -17.1 / 37.9 & 35.8 / -13.6 / 34.0  \\
PromptAttack$_{cx}$ & \underline{53.3} / \underline{-8.0} / \underline{16.3} & 54.1 / -8.5 / 16.3 & \textbf{25.0 / -31.0 / 60.0} & \underline{35.7} / \underline{-13.7} / \underline{34.1} \\
PromptAttack$_{cn}$ & \textbf{52.0 / -9.3 / 18.2} & \textbf{52.8 / -9.8 / 18.4} & \underline{37.4} / \underline{-18.6} / \underline{40.6} & 35.8 / -13.6 / 33.3 \\

\hline
\end{tabular}
\caption{Attack effectiveness results on MultiWOZ 2.1. \textbf{JGA} (\%): joint goal accuracy; \textbf{$\Delta$} (\%): the absolute difference between original JGA and JGA after attacking; \textbf{ASR} (\%): attack success rate. $\downarrow$ ($\uparrow$): denotes whether the lower (or higher) the better from an attack perspective. *: denotes the method that introduces new slot values. We highlight the \textbf{best} and the \underline{second best} results.}
\label{tab:results-2.1}
\end{table*}

\subsection{Victim Models}

We choose the \textbf{TripPy} DST~\cite{heck-etal-2020-trippy} as our base model to train our adversarial prompts since classification-based models have better performance and are more robust than generation-based models~\cite{liu-etal-2021-robustness}. Demonstrating the susceptibility of TripPy to our adversarial examples can reveal the limitations of current DSTs, but we further explore the \textit{transferability} of the prompt-based attacks. 

Transferability reflects the generalization of the attack methods, meaning that adversarial examples generated for one model can also effectively attack other models~\cite{zhang2020adversarial}. Hence, we also evaluate the prompt-based approach learned over TripPy by targeting our adversarial examples on other popular DSTs:  \textbf{TRADE}~\cite{wu2019transferable}, \textbf{SimpleTOD}~\cite{hosseini-etal-2020-simple}, and \textbf{CoCo}~\cite{li2021coco}, one of the state-of-the-art models.\footnote{These models are fine-tuned on MultiWOZ 2.1 using code from CoCo (\url{https://github.com/salesforce/coco-dst}) and follow the same post-processing strategy as CoCo. BSD 3-Clause License.} Additional information about the implementations can be found in Appendix~\ref{sec:implementation}.

\section{Experimental Results}
Given this setup, we now investigate the four experimental research questions in turn.

\subsection{Attack Effectiveness (RQ1)}
\label{ssec:effectiveness}

First, are the adversarial examples learned by PromptAttack effective? Table~\ref{tab:results-2.1} summarizes the results for three versions of PromptAttack versus the baselines for the four different DSTs (TripPy, CoCo, SimpleTOD, and TRADE). We consider the  discrete version of PromptAttack (denoted as \textbf{PromptAttack$_{d}$}) and two continuous versions: one is optimized by maximizing the training loss (denoted as \textbf{PromptAttack$_{cx}$}), while the other one is optimized by minimizing the loss (denoted as \textbf{PromptAttack$_{cn}$}).

\paragraph{Attack Performance.}
First, let's focus on the TripPy column. All versions of PromptAttack are learned over TripPy and then applied here so we can assess the susceptibility of a popular DST to adversarial examples. The four baselines lead to some degradation in terms of accuracy (JGA), with SD performing the best with a JGA of 56.5 (a 4.8 drop from the original DST).\footnote{We attribute this good attack performance since although this method maintains ground truth slot-value labels unchanged, it prepends new slot values before the original slot values in the user utterance. This operation is effective because it can easily confuse the model to decide which slot values are the truth slot values. In contrast, our prompt-based approaches are designed to make very few changes and to avoid introducing new slot values.} The three prompt-based learning approaches result in strong degradation in terms of accuracy, ranging from 7.7 to 9.3 drops relative to the original. We observe that our PromptAttack models significantly outperform SC-EDA, TP, and BERT-M, the methods without introducing new slot values in the adversarial examples, in terms of JGA and ASR. Compared with the best baseline method among these three, BERT-M, PromptAttack$_{cn}$ decreases the JGA by 6.9 and increases ASR by 13.2, respectively. In addition, for the method introducing new slot values, SD, PromptAttack$_{cn}$ outperforms it by 4.5 and 8.9. Hence, these observations reveal the attack effectiveness of our proposed PromptAttack methods over these baselines no matter whether the methods introduce new slot values or not.

\paragraph{Transferability.}
To test the transferability of the generated adversarial examples, we take the examples trained over TripPy and then use them to attack other victim models CoCo, SimpleTOD, and TRADE. For CoCo and SimpleTOD, we see that PromptAttack outperforms these four baselines. Our best method PromptAttack$_{c}$ achieves 52.8 and 25.0 JGA when attacking CoCo and SimpleTOD, showing better transferability than PromptAttack$_{d}$. For TRADE, PromptAttack$_{c}$ shows better attack performance than baselines without introducing new slot values significantly. Specifically, PromptAttack$_{cx}$ shows a decrease of 10.2 and an increase of 20.0 in terms of JGA and ASR, respectively. In general, our PromptAttack methods show good transferability: the adversarial examples generated for one victim model can also be used to attack another model effectively.

\subsection{Adversarial Example Quality (RQ2)}
\begin{table}[!bp]
\centering
\resizebox{\columnwidth}{!}{\begin{tabular}{ccccc} \hline
\textbf{Method}       & \textbf{PER$\downarrow$} & \textbf{PPL$\downarrow$} & \textbf{Semantic$\uparrow$} & \textbf{Grammar$\uparrow$} \\ \hline
Original     & -  & 173.7 & -        & 4.8    \\ \hline
SC-EDA       & \underline{13.1} & 773.8 & 2.5      & 2.7     \\
TP           & 74.4$^{\dagger}$ & 352.4 & 2.6      & \textbf{4.8}     \\
SD*          & 30.4$^{\dagger}$ & 270.4 & \textbf{4.3}      & 4.1     \\
BERT-M       & 28.1 & 221.3 & 2.8      & 4.3     \\ \hline
Adv (7.7\%)  & \textbf{7.7} & \textbf{169.0} & \textbf{4.3}      & \underline{4.4}     \\
Adv (28.1\%) & 28.1 & \underline{177.6} & \underline{3.3}      & 3.8   \\ \hline 
\end{tabular}}
\caption{Automatic evaluation and human evaluation results. \textbf{PER}: perturbation ratio; \textbf{PPL}: perplexity of generated adversarial examples representing fluency. \textbf{$\downarrow$ ($\uparrow$)} denotes whether the lower (or higher) is the better. $^{\dagger}$: results are from original papers. \textbf{*} denotes the method that introduces new slot values. Adv (*): adversarial examples from PromptAttack$_{cn}$ with different perturbation ratios which lead the victim model's accuracy to 0. We highlight the \textbf{best} and the \underline{second best} results.}
\label{tab:quality}
\end{table}
Next, we examine whether the generated adversarial examples are of good quality. First, are they fluent with a low perturbation ratio? We automatically measure the perturbation ratio (PER) between the original input and adversarial examples, and the fluency by computing the perplexity (PPL). The lower perturbation ratio represents fewer perturbed tokens in original utterances and lower perplexity indicates better fluency.
From Table~\ref{tab:quality} we observe that the PromptAttack methods achieve low perplexity and show good fluency with quite a low perturbation ratio. Specifically, our method PromptAttack$_{cn}$ (7.7\%) achieves 169.0 PPL, showing better fluency than PromptAttack$_{cn}$ (28.1\%) and baselines. Although SC-EDA has a lower perturbation ratio than our PromptAttack$_{cn}$ (28.1\%), it shows less attack effectiveness (Section~\ref{ssec:effectiveness}) and worse fluency. Thus, there are trade-offs between perturbation ratio and attack effectiveness.

Second, do the adversarial examples preserve the semantics of the un-perturbed original sentences? That is, does an utterance asking for a cheap restaurant lead to an adversarial example that also asks for a cheap restaurant though tricking the DST to output expensive? To answer this question, we conduct a human evaluation on semantics preservation and grammatical correctness. We first shuffled 150 examples: 50 original un-perturbed sentences, 50 adversarial examples with a 7.7\% perturbation ratio, and 50 with a 28.1\% perturbation ratio (following the analysis in Section~\ref{ssec:ratio}). For the adversarial examples, each attacks the victim model successfully leading to an accuracy of 0. Following~\cite{jin-etal-2020-robust,li-etal-2020-bert-attack}, we ask three human judges to rate how well a randomly chosen sentence preserves the semantics of the original sentence (\textit{semantic}), how grammatically correct the sentence is (\textit{grammar}), on a scale from 1 to 5. We report the average score across the three judges in Table~\ref{tab:quality}.

As we can see, the semantic score and grammar score of the adversarial examples are close to the original ones.
We find that when the perturbation is reasonable (around 8\%), the semantics of the original sentence are preserved quite well (scoring 4.3 for adversarial examples). Further, the grammatical quality of the sentence is also maintained well (4.8 versus 4.4). Even as the perturbation ratio increases to approximately 28\%, our approach continues to uphold good semantic preservation (3.3) while retaining satisfactory grammar quality (3.8). Overall, our method consistently generates high-quality adversarial examples by effectively preserving semantics, maintaining grammatical quality and fluency, and keeping a low perturbation ratio.

\subsection{Impact of PromptAttack Design (RQ3)} 
We now explore the impact of different settings on our proposed methods.

\subsubsection{Ratio of Perturbed Tokens}
\label{ssec:ratio}

First, our prompt-based approach can control how many tokens we want to change in the user utterances, which gives it flexibility. Since the perturbation ratio represents the semantic consistency between the original examples and adversarial examples and there are trade-offs between the attack effectiveness and perturbation ratio, it is important to investigate the influence of the ratio of perturbed tokens on attacking ability.

We take $\max (1, perturbation\_ratio*l_{t})$ as the number of perturbed tokens, where $l_{t}$ denotes the length of pre-processed utterances. We set the perturbation ratio of tokens that we could perturb to 10\%, 30\%, 50\%, 80\%, and 100\%, that is 7.7\%, 10.2\%, 15.2\%, 22.6\%, and 28.1\% of the average length of all input examples. More data analysis can be found in Appendix~\ref{apx:data-analysis}. 

Table~\ref{tab:max-min} shows the evaluation of attack performance and fluency of generated adversarial examples from PromptAttack$_{cx}$ and PromptAttack$_{cn}$. We observe that for these two methods, the more tokens we perturb, the lower JGA and higher ASR we get, showing better attack ability, which is consistent with our intuition. Thus, as the ratio of perturbed tokens increases, our proposed method PromptAttack achieves better attack performance while maintaining good fluency.
\begin{table}[!tp]
\centering\small
\resizebox{\columnwidth}{!}{\begin{tabular}{cc|ccccc}\hline
\multicolumn{2}{l|}{\multirow{2}{*}{}} & \textbf{7.7\%} & \textbf{10.2\%} & \textbf{15.2\%} & \textbf{22.6\%} & \textbf{28.1\%} \\
\multicolumn{2}{l|}{} & (1.0) & (1.5) & (2.3) & (3.5) & (4.4) \\ \hline
\multirow{3}{*}{P$_{cx}$} & JGA$\downarrow$ &59.0 &58.1 &56.6 &55.1 & 53.3\\
                     & ASR$\uparrow$ & 4.6 &6.3  &9.5  &12.8 & 16.3\\
                     & PPL$\downarrow$ &159.4&155.9&157.5&167.0&175.5\\\hline
\multirow{3}{*}{P$_{cn}$} & JGA$\downarrow$ &58.9 &58.1 &56.4 & 54.0& 52.0\\
                     & ASR$\uparrow$ & 4.9 &6.2  &9.7  & 14.4& 18.2\\
                     & PPL$\downarrow$ &169.0&173.7&177.1&172.2&177.6\\\hline
\end{tabular}}
\caption{Results of PromptAttack$_{cx}$ (P$_{cx}$) and PromptAttack$_{cn}$ (P$_{cn}$) with different perturbation ratio. (*) denotes the average perturbed token numbers.}
\label{tab:max-min}
\end{table}

\begin{table}[!bp]
\centering\small
\resizebox{\columnwidth}{!}{\begin{tabular}{ccccccc} \hline
\multirow{2}{*}{} & \multicolumn{2}{c}{\textbf{P$_{5}$}} & \multicolumn{2}{c}{\textbf{P$_{10}$}} & \multicolumn{2}{c}{\textbf{P$_{15}$}}\\
                  & JGA$\downarrow$ &  ASR$\uparrow$ & JGA$\downarrow$ &  ASR$\uparrow$ & JGA$\downarrow$ &  ASR$\uparrow$  \\\hline
7.7\% & 59.0 & 4.6 & 59.2 & 4.3 & 59.3 & 4.6 \\
10.2\% & 58.1 & 6.3 & 58.5 & 5.9 & 58.5 & 6.0 \\
15.2\% & 56.6 & 9.5 & 57.0 & 8.8 & 57.2 & 8.8 \\
22.6\% & 55.1 & 12.8 & 55.3 & 12.6 & 55.7 & 12.3 \\
28.1\% & 53.3 & 16.3 & 53.5 & 15.9 & 54.1 & 15.1 \\
\hline
\end{tabular}}
\caption{Results of PromptAttack$_{cx}$ with different prompts length and perturbation ratios. \textbf{P}$_{*}$ denotes the prompt length.}
\label{tab:prompt-length}
\end{table}
\subsubsection{Prompt Length}

Next, we explore the effect of different continuous prompt lengths. Shorter prompts have fewer tunable parameters, which means under the same training setting, it would be faster to optimize and find the most effective adversarial prompts. We train continuous prompts with different length: 5 tokens, 10 tokens, and 15 tokens using PromptAttack$_{cx}$. Table~\ref{tab:prompt-length} shows that under different prompt lengths, with the increase of perturbation ratio, the model achieves better attack performance. Under the same perturbation ratios, the model with 5-token prompt achieves modest lower JGA and higher ASR. For example, when the perturbation ratio is 28.1\%, PromptAttack$_{cx}$ with 5-token prompt gains lower JGA than PromptAttack$_{cx}$ with 10-token prompt and PromptAttack$_{cx}$ with 15-token prompt by 0.2 and 0.8, respectively, and higher ASR by 0.4 and 1.2, indicating slightly better attack performance.

\subsection{Defense against Attack (RQ4)} 
\label{ssec:defense}
Finally, we turn to the challenge of defending a DST in the presence of such adversarial examples. We aim to answer two questions: i) can our generated adversarial examples be used to improve the performance of current DST models? and ii) can our attack method bypass such a defense method?

One of the most effective approaches to increase the robustness of a model is adversarial training, which injects adversarial examples into the training data to increase model robustness intrinsically~\cite{bai2021recent}. Specifically, we first apply our attack methods on the original training dataset to generate adversarial examples. Then we re-train the TripPy model on the training set augmented by the adversarial training examples and evaluate the performance on original test set. As shown in Table~\ref{tab:defense}, the new defended DST model improves JGA on the original test set from 61.3 to 67.3 by 6.0, which outperforms results reported by the state-of-the-art DST model CoCo (62.6) by 4.7. This encouraging result shows that adversarial examples from our attack method can be a good source for data augmentation.

To evaluate the robustness of such an augmented DST model against our proposed attack methods, we next test how well our adversarial examples perform. From Table~\ref{tab:defense} we observe that the attack methods still show strong attack ability on the new DST model. Thus, there is an opportunity to explore stronger defense methods to strengthen DSTs against such prompt-based attacks.
\begin{table}[!tp]
\centering\small
\begin{tabular}{ccccc} \hline
 & \textbf{JGA$_d$} & \textbf{JGA$_o$} & \textbf{ASR$_d$} & \textbf{ASR$_o$} \\ \hline
Original         & 67.3       & 61.3       & -          & -           \\\hline
SC-EDA            & 66.5       & 60.5       & 1.8        & 1.9        \\
TP           & 65.9       & 60.3       & 5.5        & 5.6        \\
SD$^{*}$          & 61.4       & 56.5       & 10.1       & 9.3       \\
BERT-M            & 64.5       & 58.9       & 5.0        & 5.0       \\\hline
PromptAttack$_d$          & 60.0       & 55.8       & 12.6       & 11.3      \\
PromptAttack$_{cx}$        & 58.3       & 53.3       & 16.3       & 16.3      \\
PromptAttack$_{cn}$        & \textbf{56.8}       & \textbf{52.0}       & \textbf{18.5}       & \textbf{18.2}      \\\hline
\end{tabular}
\caption{Defense results. d: defended DST model; o: original DST model.}
\label{tab:defense}
\end{table}

\section{Conclusion}
In this paper, we present a prompt-based learning approach that can generate effective adversarial examples for probing DST models. Through experiments over four state-of-the-art DSTs, our framework achieves the greatest reduction in accuracy with the best attack success rate. Moreover, the generated adversarial examples maintain good fluency and low perturbation ratio, evidence that they are close to legitimate non-adversarial user inputs. We also show our generated adversarial examples can bolster a DST by augmenting the original training data with adversarial examples. We find that both discrete and continuous adversarial prompts are capable of generating effective adversarial examples. Discrete prompts are more interpretable while continuous prompting allows us to search for optimal adversarial prompts more efficiently, and generates more effective adversarial examples.

\section*{Limitations}
The natural idea to improve robustness is to add adversarial examples to the training set and retrain the model. However, generating adversarial examples for a large training set can be very time-consuming. Thus, it would be interesting to explore more efficient methods that implicitly involved adversarial examples in the training process, e.g.,~\cite{yang2022prompting}. 

\section*{Ethics Statement}
The proposed methods could also be applied to natural language generation tasks, like dialogue response generation. The misuse of such methods may generate biased or offensive responses.

\section*{Acknowledgements}
We appreciate the authors of the CoCo paper for making their code accessible to the public and for taking the time to address our inquiries.

\bibliography{anthology,custom}
\bibliographystyle{acl_natbib}

\appendix

\section{Implementation Details}
\label{sec:implementation}
\subsection{Constructing Adversarial Examples}
We use the TripPy DST model~\cite{heck-etal-2020-trippy} as the victim model, which uses the 12-layer pre-trained BERT-base-uncased model~\cite{devlin2019bert} as the context encoder and has 768 hidden units, 12 self-attention heads, and 110M parameters. We train our adversarial prompts for 10 epochs with an initial learning rate of $1\times 10^{-4}$. The LR decay linearly with a warmup proportion of 0.1. We use Adam optimizer for optimization and set the maximum input sequence length of user utterance $l_{u}$ to 180 and the number of prompt tokens $l_{p}$ to $\{5, 10, 15\}$. The total length of the input is $l_{u}+l_{p}$. The training batch size is 64 and the evaluation batch size is 1. We evaluate the checkpoint of the prompt for each epoch and choose the one that leads to the lowest JGA on the validation set as our final adversarial prompt.
The MLM model used to generate the adversarial examples via mask-and-filling is also BERT-base-uncased.
\subsection{Training Defense Models}
We train the TripPy defense DST model on the training dataset augmented with adversarial examples following the training setting in~\cite{heck-etal-2020-trippy}. The model uses the pre-trained BERT-base-uncased transformer as the context encoder front-end, which has 12 hidden layers with 768 hidden units and 12 self-attention heads each~\cite{heck-etal-2020-trippy}. The initial learning rate is set to $1\times 10^{-4}$ with a warmup proportion of 0.1 and let the LR decay linearly after the warmup phase. We use Adam optimizer for optimization and dropout on the BERT output with a rate of 30\%. The training batch size is 48 and the model is trained for 10 epochs with early stopping employed based on the JGA on the validation set. The experiments are run on 2 NVIDIA TITAN Xp GPUs.

\section{Data Analysis}
\label{apx:data-analysis}

\begin{figure}[htbp!]
\centering
    \begin{tikzpicture}[scale=0.65]
\begin{axis}[
    width=1.5\linewidth,
    height=\axisdefaultheight,
	symbolic x coords={0-4, 5-9, 10-14, 15-19, 20-24, 25-29, 30-34, 35-39},
	ylabel=Percentage(\%),
	xlabel=Number of tokens,
	xticklabel style={rotate=30},
	ybar,
	bar width=4.5pt,
	ymajorgrids=true,
    grid style=dashed,
    axis lines*=left,
    ymin=0.0, ymax=60,
    nodes near coords,
    nodes near coords align={vertical},
    every axis plot/.append style={fill},
	]
\addplot
    coordinates{(0-4,0.5) (5-9,16.5) (10-14,35.1) (15-19,25.0) (20-24,13.9) (25-29,6.0) (30-34,2.4) (35-39,0.6)};
\addplot
    coordinates{(0-4,59.3) (5-9,36.2) (10-14,4.4) (15-19,0.1)};
\addplot
    coordinates{(0-4,3.5) (5-9,38.5) (10-14,38.1) (15-19,15.1) (20-24,4.3) (25-29,0.5)};
\legend{l$_o$, l$_t$, $\Delta$}

\end{axis}
\end{tikzpicture}
    \vspace{-1cm}
\caption{Data analysis. \textbf{l$_{o}$}: original length of user utterance; \textbf{l$_{t}$}: after data pre-processing, the number of available tokens that can be perturbed; \textbf{$\Delta$}: $l_{o}-l_{t}$.}    
\label{fig:data-analysis}
\end{figure}
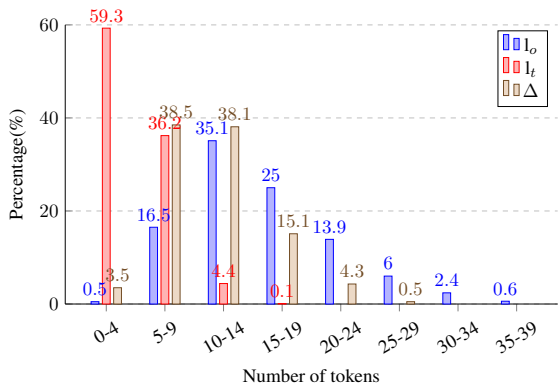

Figure~\ref{fig:data-analysis} shows the distribution of length of original user utterance $l_{o}$, the length of utterances after removing stop words, slot and value related tokens $l_{t}$, and the difference between them, that is $\Delta=l_{o}-l_{t}$. We can see, 95.5\% of user utterances have fewer than 10 tokens that could be perturbed and 59.3\% of them could perturb less than 4 tokens.

\section{Success Generation Rates of Adversarial Examples}
To assess whether there are any masked tokens kept unchanged after the adversarial example construction process, we evaluate the success generation rates of adversarial examples generated using our methods. we observe that PromptAttack$_d$ achieves a success rate of 0.94, PromptAttack$_{cx}$ achieves 0.95, and PromptAttack$_{cn}$ achieves 0.96. These findings indicate that our methods demonstrate a high rate of successful generation. And it is important to note that unchanged utterances will not attack the models successfully.

\section{Data Augmentation}
In Section~\ref{ssec:defense}, TripPy trained on original training data augmented with our generated adversarial examples improves TripPy by 6.0\% and outperforms CoCo by 4.7\% when evaluated on the original test set following the same post-processing strategy as CoCo~\cite{li2021coco}. When using TripPy's default cleaning, the comparison results with previous methods are shown in Table~\ref{tab:sota}.
\begin{table}[htbp!]
\centering\small
\begin{tabular}{lc} \hline
\textbf{Model} & \textbf{JGA (\%)} \\ \hline
TRADE~\cite{wu2019transferable} & 46.00$^{\dagger}$ \\
TripPy~\cite{heck-etal-2020-trippy} & 55.29$^{\dagger}$ \\
SimpleTOD~\cite{hosseini-etal-2020-simple} & 55.76$^{\dagger}$ \\
ConvBERT-DG + Multi~\cite{mehri2020dialoglue} & 58.70$^{\dagger}$ \\
TripPy + SCoRE~\cite{yu2021score} & 60.48$^{\dagger}$ \\
TripPy + CoCo~\cite{li2021coco} & 60.53$^{\dagger}$ \\
Ours & 60.56 \\
TripPy + SaCLog~\cite{dai-etal-2021-preview} & 60.61$^{\dagger}$ \\ \hline

\end{tabular}
\caption{DST results on MultiWOZ 2.1. $^{\dagger}$ denotes that results are from original papers.}
\label{tab:sota}
\end{table}

\end{document}